\documentclass{article}
\usepackage[final]{nips_2016} 
\usepackage[utf8]{inputenc}
\usepackage{natbib}
\usepackage{color}
\usepackage{amsmath}
\usepackage{amssymb}
\usepackage{graphicx} 
\usepackage{caption}
\usepackage{subcaption}
\usepackage{graphicx}
\usepackage{epsfig}
\usepackage{bm}
\usepackage{hyperref}
\usepackage{cleveref}

\newcommand{\bx}{\mathbf{x}}

\newcommand{\bz}{\mathbf{z}}

\title{Instance Normalization:\\ The Missing Ingredient for Fast Stylization}

\author{
  Dmitry Ulyanov \\
  Computer Vision Group \\
  Skoltech \& Yandex \\  
  Russia \\
  \texttt{dmitry.ulyanov@skoltech.ru} \\
  \And
  Andrea Vedaldi \\
  Visual Geometry Group \\ 
  University of Oxford \\ 
  United Kingdom \\
  \texttt{vedaldi@robots.ox.ac.uk} \\
   \AND
  Victor Lempitsky \\
  Computer Vision Group \\
  Skoltech \\  
  Russia \\
  \texttt{lempitsky@skoltech.ru} \\
}
  
\begin{document}
\maketitle
\begin{abstract}
It this paper we revisit the fast stylization method introduced in \cite{DBLP:conf/icml/UlyanovLVL16}. We show how a small change in the stylization architecture results in a significant qualitative improvement in the generated images. The change is limited to swapping batch normalization with instance normalization, and to apply the latter both at training and testing times. The resulting method can be used to train high-performance architectures for real-time image generation. The code is available at \url{https://github.com/DmitryUlyanov/texture_nets}. Full paper can be found at \url{https://arxiv.org/abs/1701.02096}.
\end{abstract}

\section{Introduction}\label{s:intro}

\begin{figure}
    \centering
   \includegraphics[width=\textwidth]{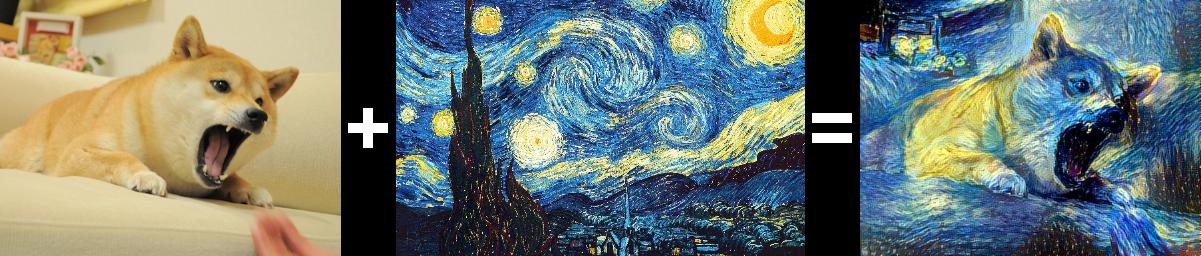}
\caption{Artistic style transfer example of \cite{Gatys_2016_CVPR} method. }\label{fig:doge}
\end{figure}

The recent work of \cite{Gatys_2016_CVPR} introduced a method for transferring a style from an image onto another one, as demonstrated in \cref{fig:doge}. The stylized image matches simultaneously selected statistics of the style image and of the content image. Both style and content statistics are obtained from a deep convolutional network pre-trained for image classification. The style statistics are extracted from shallower layers and averaged across spatial locations whereas the content statistics are extracted form deeper layers and preserve spatial information. In this manner, the style statistics capture the ``texture'' of the style image whereas the content statistics capture the ``structure'' of the content image.

Although the method of  Gatys~et.~al produces remarkably good results, it is computationally inefficient. The stylized image is, in fact, obtained by iterative optimization until it matches the desired statistics. In practice, it takes several minutes to stylize an image of size $512\times 512$. Two recent works, \cite{DBLP:conf/icml/UlyanovLVL16} \cite{DBLP:journals/corr/JohnsonAL16}, sought to address this problem by \emph{learning equivalent feed-forward generator networks} that can generate the stylized image in a single pass. These two methods differ mainly by the details of the generator architecture and produce results of a comparable quality; however, neither achieved as good results as the slower optimization-based method of Gatys~et.~al.

In this paper we revisit the method for feed-forward stylization of~\cite{DBLP:conf/icml/UlyanovLVL16} and show that a small change in a generator architecture leads to much improved results. The results are in fact of comparable quality as the slow optimization method of Gatys~et al. but can be obtained in real time on standard GPU hardware. The key idea (\cref{s:method}) is to replace batch normalization layers in the generator architecture with instance normalization layers, and to keep them at test time (as opposed to freeze and simplify them out as done for batch normalization). Intuitively, the normalization process allows to remove instance-specific contrast information from the content image, which simplifies generation. In practice, this results in vastly improved images (\cref{s:experiments}).

\begin{figure}
    \centering
    \begin{subfigure}[b]{0.3\textwidth}
        \includegraphics[width=\textwidth]{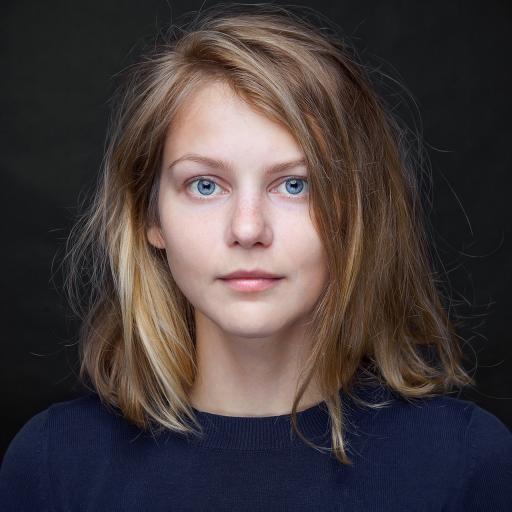}
        \caption{Content image.}
    \end{subfigure}
    \begin{subfigure}[b]{0.3\textwidth}
        \includegraphics[width=\textwidth]{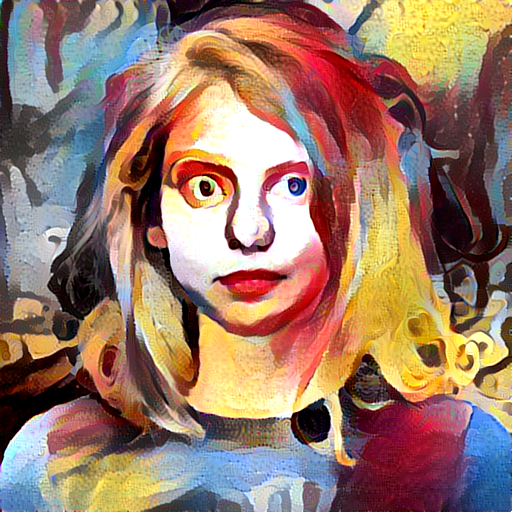}
        \caption{Stylized image.}
    \end{subfigure}
    \\
    \centering
    \begin{subfigure}[b]{0.3\textwidth}
        \includegraphics[width=\textwidth]{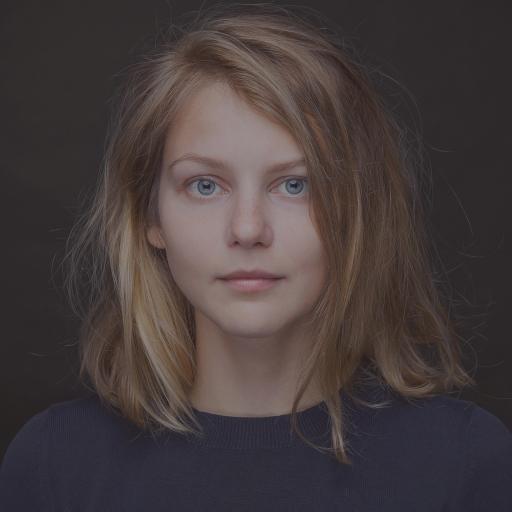}
        \caption{Low contrast content image.}
    \end{subfigure}
    \begin{subfigure}[b]{0.3\textwidth}
        \includegraphics[width=\textwidth]{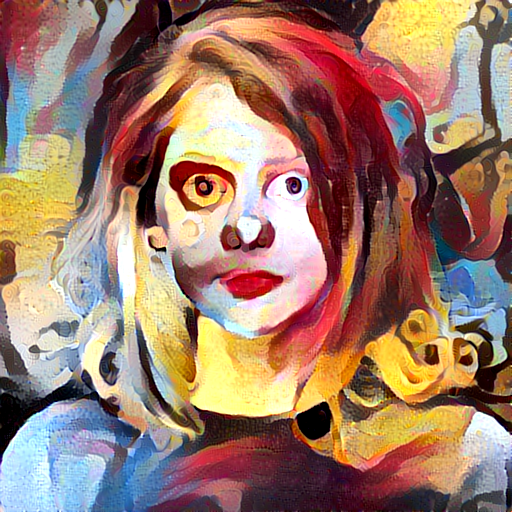}
        \caption{Stylized low contrast image.}
    \end{subfigure}
\caption{A contrast of a stylized image is mostly determined by a contrast of a style image and  almost independent of a content image contrast. The stylization is performed with method of \cite{Gatys_2016_CVPR}.}\label{fig:contrast}
\end{figure}

\begin{figure}
    \centering
    \begin{subfigure}[b]{0.31\textwidth}
        \includegraphics[width=\textwidth]{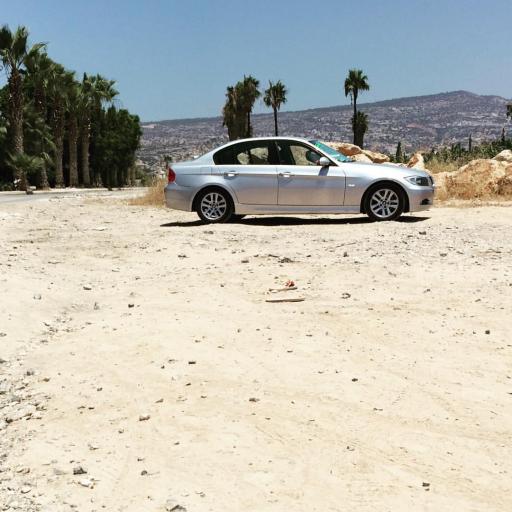}
    \end{subfigure}
    \begin{subfigure}[b]{0.31\textwidth}
        \includegraphics[width=\textwidth]{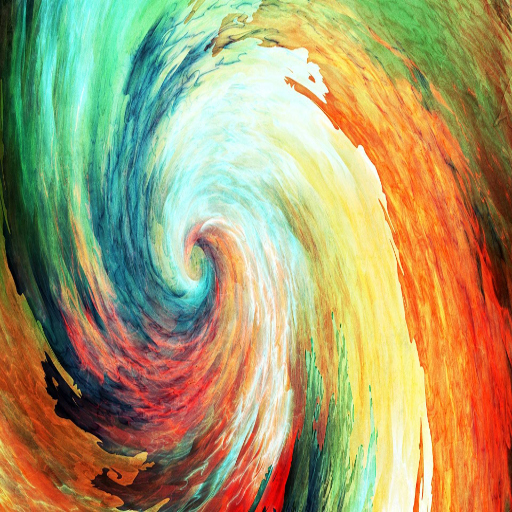}
    \end{subfigure}
    \begin{subfigure}[b]{0.31\textwidth}
        \includegraphics[width=\textwidth]{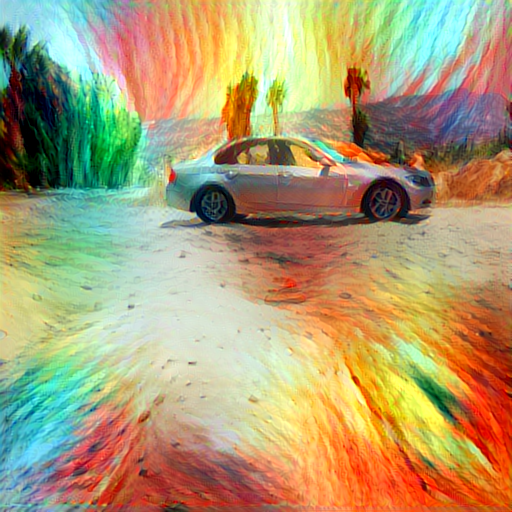}
    \end{subfigure}
    \\
    \begin{subfigure}[b]{0.31\textwidth}
        \includegraphics[width=\textwidth]{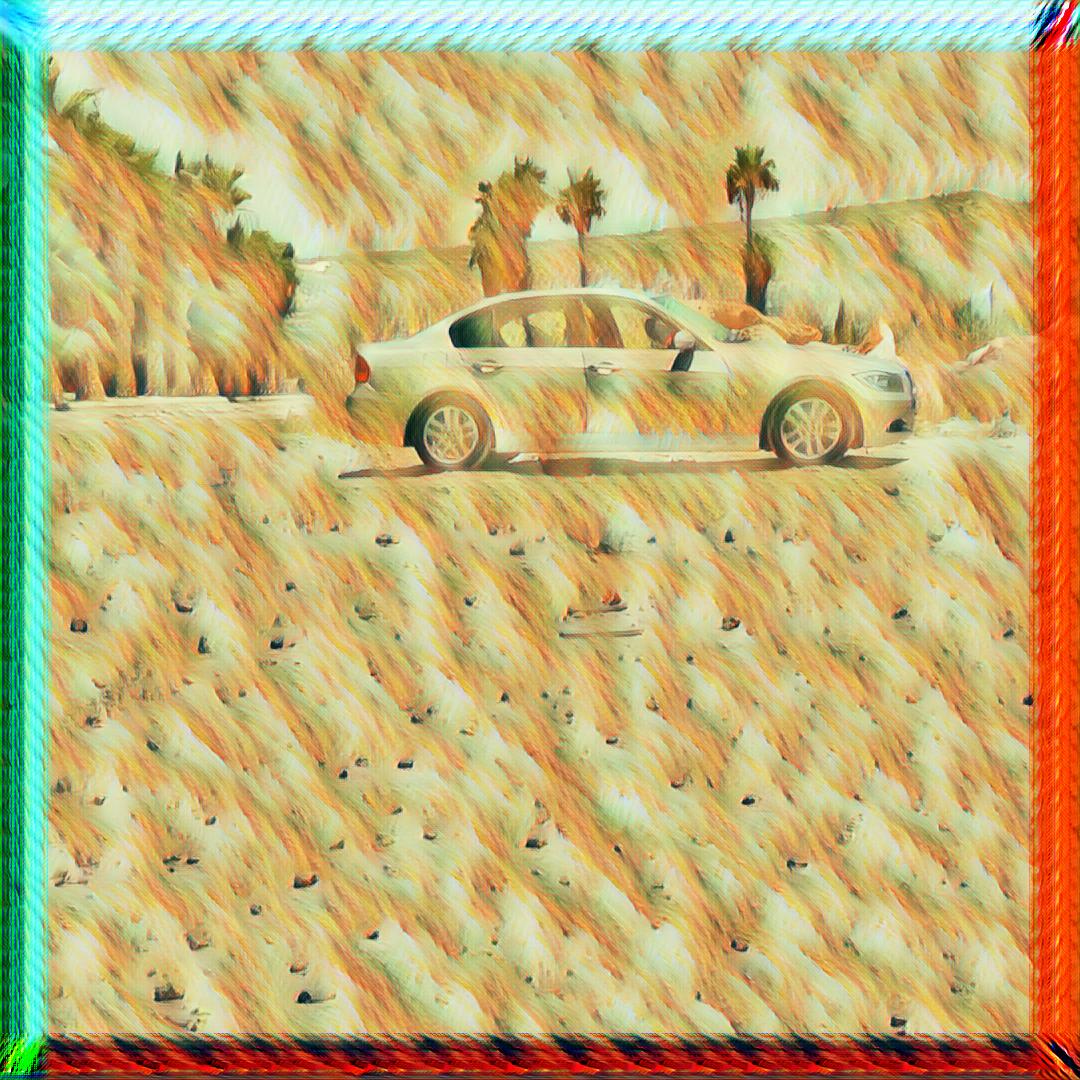}
    \end{subfigure}
    \begin{subfigure}[b]{0.31\textwidth}
        \includegraphics[width=\textwidth]{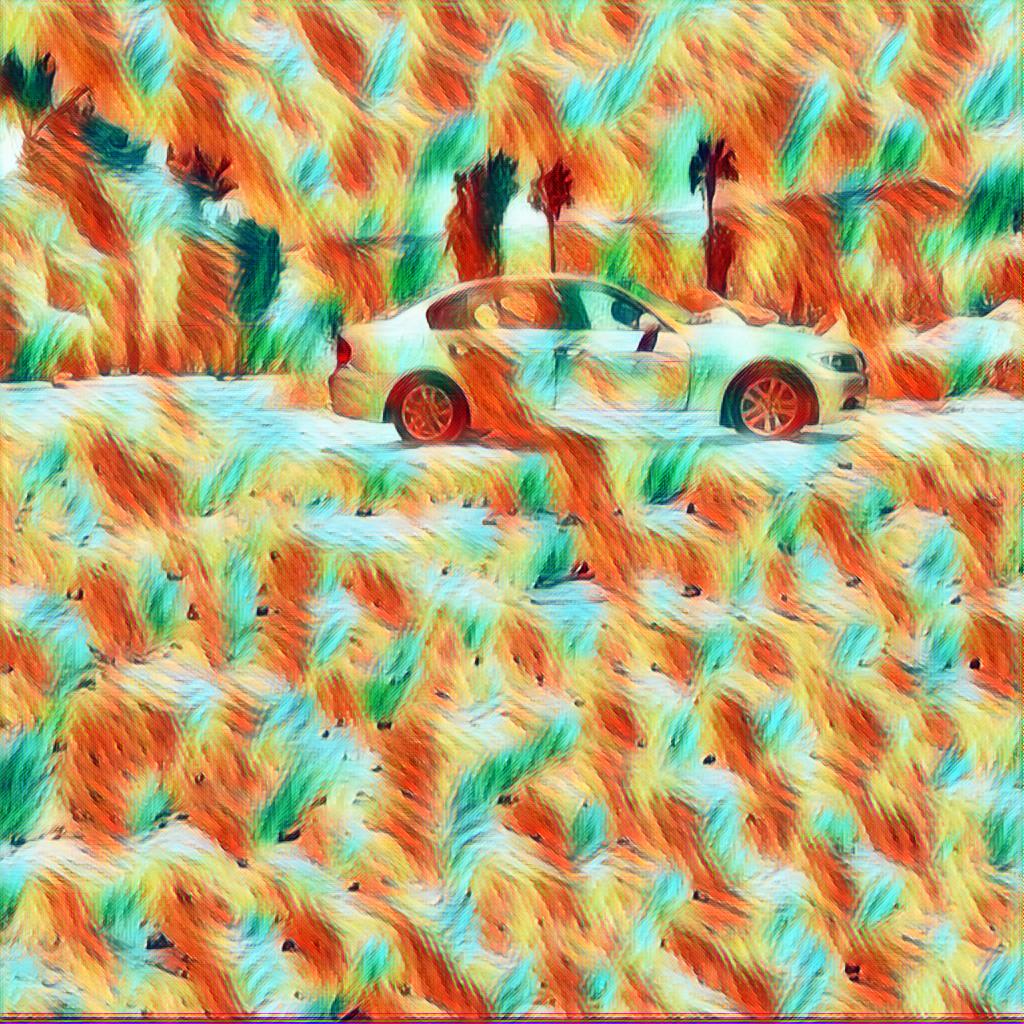}
    \end{subfigure}
    \begin{subfigure}[b]{0.31\textwidth}
        \includegraphics[width=\textwidth]{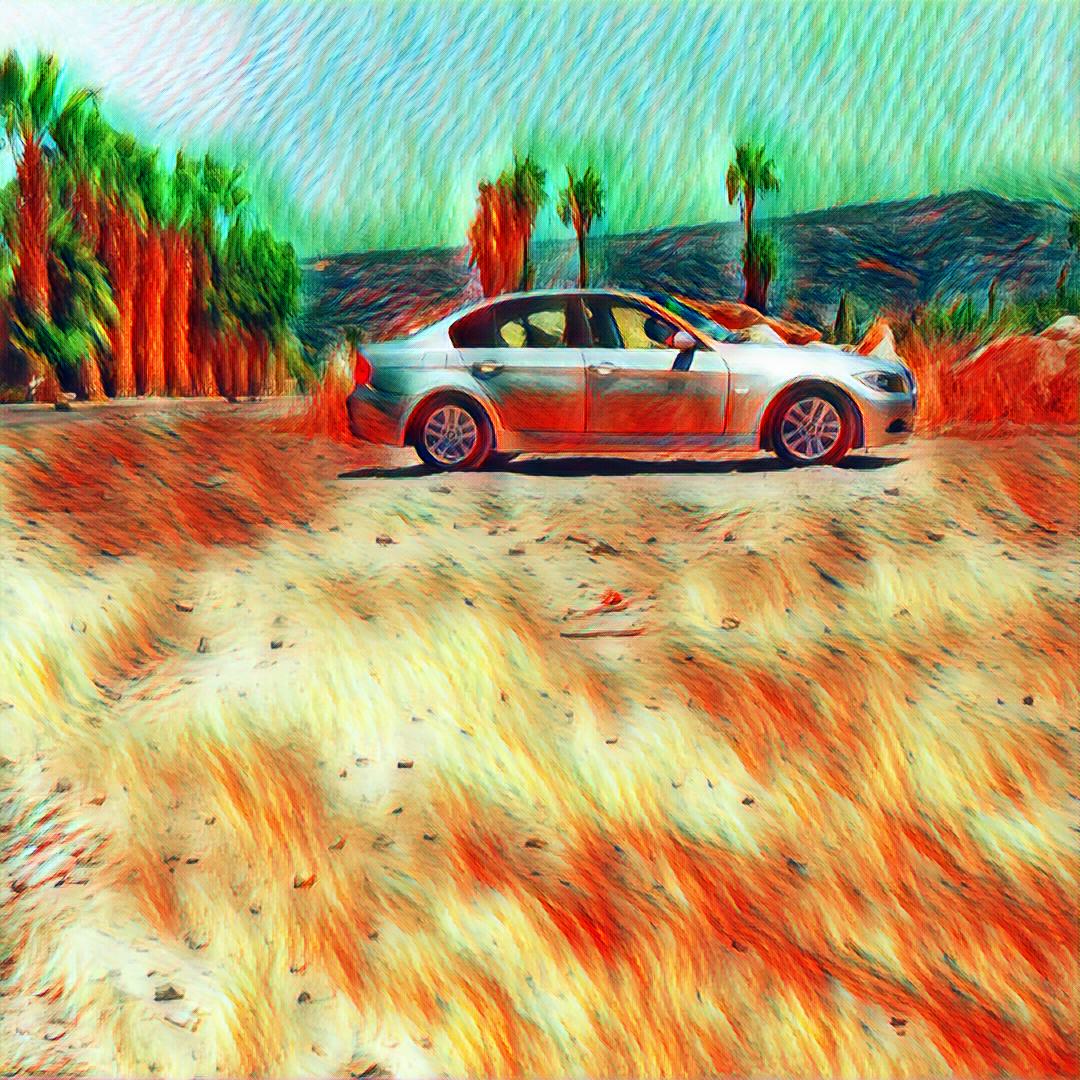}
    \end{subfigure}
\caption{Row 1: content image (left), style image (middle) and style transfer using method of Gatys~et.~al (right). Row 2: typical stylization results when trained for a large number of iterations using fast stylization method from \cite{DBLP:conf/icml/UlyanovLVL16}: with zero padding (left), with a better padding technique (middle), with zero padding and instance normalization (right).} \label{fig:comparison}
\end{figure}

\section{Method}\label{s:method}
    
The work of~\cite{DBLP:conf/icml/UlyanovLVL16} showed that it is possible to learn a generator network $g(\bx,\bz)$ that can apply to a given input image $\bx$ the style of another $\bx_0$, reproducing to some extent the results of the optimization method of Gatys~et al. Here, the style image $\bx_0$ is fixed and the generator $g$ is learned to apply the style to any input image $\bx$. The variable $\bz$ is a random seed that can be used to obtain sample stylization results.

The function $g$ is a convolutional neural network learned from examples. Here an example is just a content image $\bx_t,t=1,\dots,n$ and learning solves the problem
$$
\min_{g} \frac{1}{n} \sum_{t=1}^n
\mathcal{L}(\bx_0,\bx_t,g(\bx_t,\bz_t))
$$
where $\bz_t \sim \mathcal{N}(0,1)$ are i.i.d. samples from a Gaussian distribution. The loss $\mathcal{L}$ uses a pre-trained CNN (not shown) to extracts features from the style $\bx_0$ image, the content image $\bx_t$, and the stylized image $g(\bx_t,\bz_t)$, and compares their statistics as explained before.

While the generator network $g$ is fast, the authors of~\cite{DBLP:conf/icml/UlyanovLVL16} observed that learning it from too many training examples yield poorer \emph{qualitative} results. In particular, a network trained on just 16 example images produced better results than one trained from thousands of those. The most serious artifacts were found along the border of the image due to the zero padding added before every convolution operation (see \cref{fig:comparison}). Even by using more complex padding techniques it was not possible to solve this issue. Ultimately, the best results presented in \cite{DBLP:conf/icml/UlyanovLVL16} were obtained using a small number of training images and stopping the learning process early. We conjectured that the training objective was too hard to learn for a standard neural network architecture.
    
A simple observation is that the result of stylization should not, in general, depend on the contrast of the content image (see \cref{fig:contrast}). In fact, the \textit{style loss} is designed to transfer elements from a style image to the content image such that the contrast of the stylized image is similar to the contrast of the style image. Thus, the generator network should discard contrast information in the content image. The question is whether contrast normalization can be implemented efficiently by combining standard CNN building blocks or whether, instead, is best implemented directly in the architecture.

The generators used in \cite{DBLP:conf/icml/UlyanovLVL16} and \cite{DBLP:journals/corr/JohnsonAL16} use convolution, pooling, upsampling, and batch normalization. In practice, it may be difficult to learn a highly nonlinear contrast normalization function as a combination of such layers. To see why, let $x\in\mathbb{R}^{T \times C \times W \times H}$ be an input tensor containing a batch of $T$ images. Let $x_{tijk}$ denote its $tijk$-th element, where $k$ and $j$ span spatial dimensions, $i$ is the feature channel (color channel if the input is an RGB image), and $t$ is the index of the image in the batch. Then a simple version of contrast normalization is given by: 
\begin{equation}\label{eq:contrast}
    y_{tijk} =  \frac{x_{tijk}}{\sum_{l=1}^W \sum_{m=1}^H x_{tilm}}.
\end{equation}
It is unclear how such as function could be implemented as a sequence of ReLU and convolution operator.

On the other hand, the generator network of~\cite{DBLP:conf/icml/UlyanovLVL16} does contain a normalization layers, and precisely \emph{batch normalization} ones. The key difference between \cref{eq:contrast} and batch normalization is that the latter applies the normalization to a  whole batch of images instead for single ones:
\begin{equation}\label{eq:bnorm}
    y_{tijk} =  \frac{x_{tijk} - \mu_{i}}{\sqrt{\sigma_i^2 + \epsilon}},
    \quad
    \mu_i = \frac{1}{HWT}\sum_{t=1}^T\sum_{l=1}^W \sum_{m=1}^H x_{tilm},
    \quad
    \sigma_i^2 = \frac{1}{HWT}\sum_{t=1}^T\sum_{l=1}^W \sum_{m=1}^H (x_{tilm} - mu_i)^2.
\end{equation}
In order to combine the effects of instance-specific normalization and batch normalization, we propose to replace the latter by the \emph{instance normalization} (also known as ``contrast normalization'') layer:
\begin{equation}\label{eq:inorm}
    y_{tijk} =  \frac{x_{tijk} - \mu_{ti}}{\sqrt{\sigma_{ti}^2 + \epsilon}},
    \quad
    \mu_{ti} = \frac{1}{HW}\sum_{l=1}^W \sum_{m=1}^H x_{tilm},
    \quad
    \sigma_{ti}^2 = \frac{1}{HW}\sum_{l=1}^W \sum_{m=1}^H (x_{tilm} - mu_{ti})^2.
\end{equation}
We replace batch normalization with instance normalization everywhere in  the generator network $g$. This prevents instance-specific mean and covariance shift simplifying the learning process. Differently from batch normalization, furthermore, the instance normalization layer is applied at test time as well. 

\section{Experiments}\label{s:experiments}

\begin{figure}
    \centering
    \begin{subfigure}[b]{0.45\textwidth}
        \includegraphics[width=\textwidth]{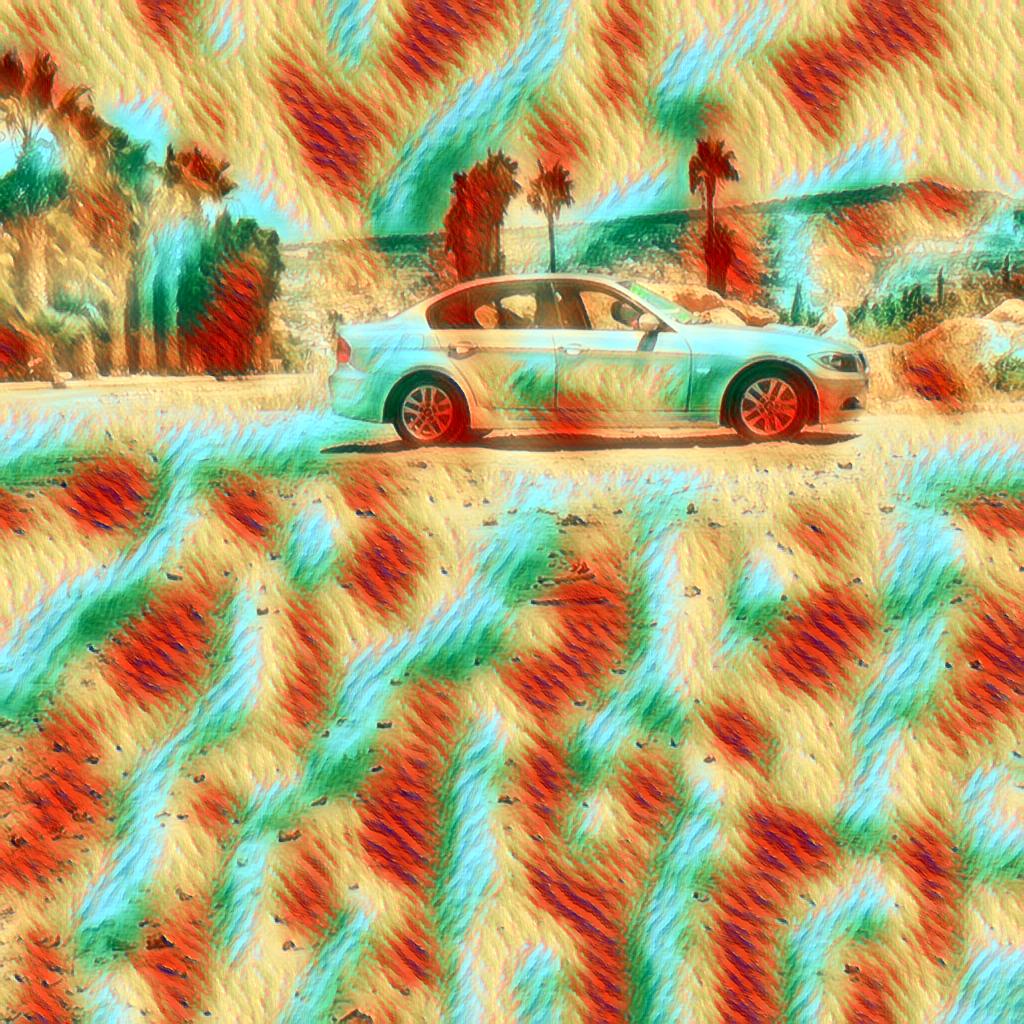}
      \end{subfigure}
    \begin{subfigure}[b]{0.45\textwidth}
        \includegraphics[width=\textwidth]{figures/car_johnson_old.jpg}
    \end{subfigure}
    \\
    \begin{subfigure}[b]{0.45\textwidth}
        \includegraphics[width=\textwidth]{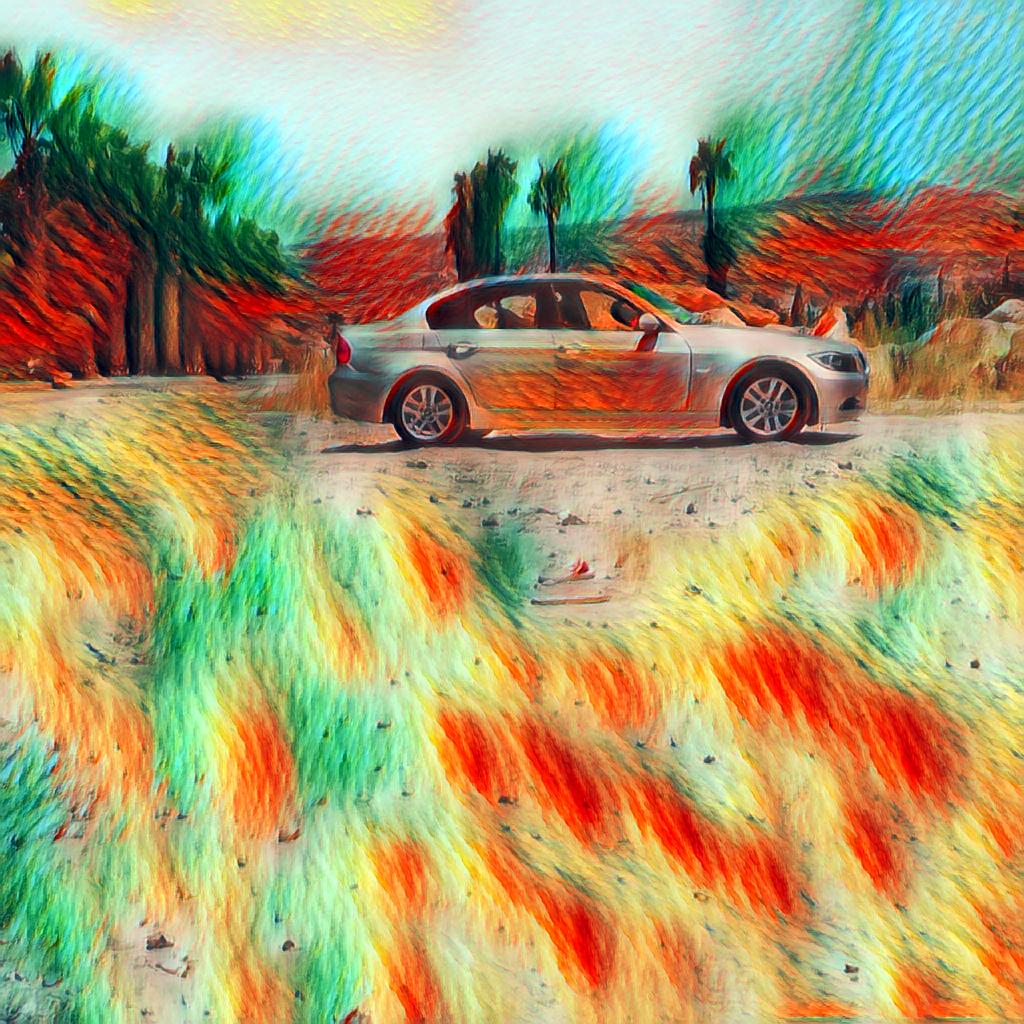}
      \end{subfigure}
    \begin{subfigure}[b]{0.45\textwidth}
        \includegraphics[width=\textwidth]{figures/car_johnson.jpg}
    \end{subfigure}
\caption{Qualitative comparison of generators proposed in \cite{DBLP:conf/icml/UlyanovLVL16} (left), \cite{DBLP:journals/corr/JohnsonAL16} (right) with batch normalization (first row) and instance normalization (second row). Both architectures benefit from instance normalization.\label{fig:before_after}} 
\end{figure}

In this section, we evaluate the effect of the modification proposed in~\cref{s:method} and replace batch normalization with instance normalization. We tested both generator architectures described in~\cite{DBLP:conf/icml/UlyanovLVL16} and~\cite{DBLP:journals/corr/JohnsonAL16} in order too see whether the modification applies to different architectures. While we did not have access to the original network by~\cite{DBLP:journals/corr/JohnsonAL16}, we carefully reproduced their model from the description in the paper. Ultimately, we found that both generator networks have similar performance and shortcomings (\cref{fig:before_after} first row).

\begin{figure}
    \centering
    \hspace{0.30\textwidth}
    \begin{subfigure}[b]{0.30\textwidth}
        \includegraphics[width=\textwidth]{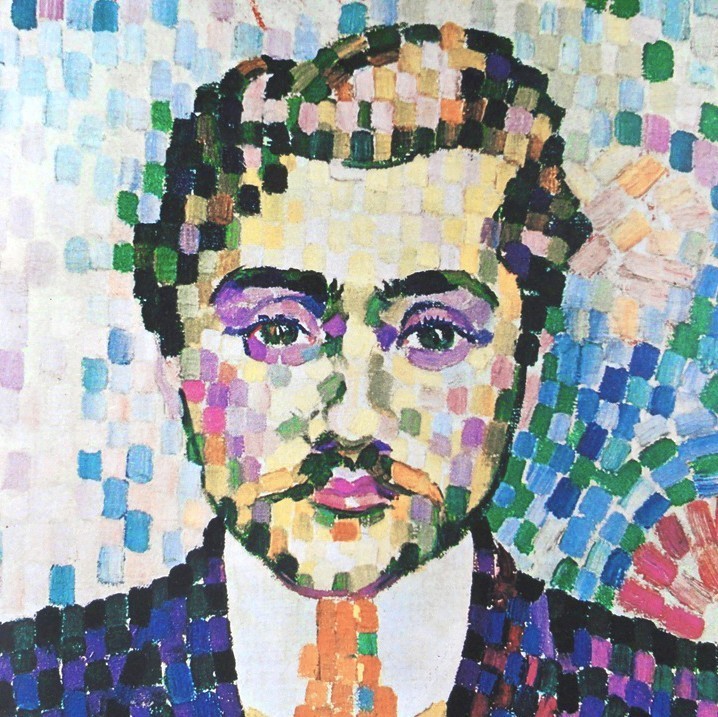}
    \end{subfigure}
    \begin{subfigure}[b]{0.30\textwidth}
        \includegraphics[width=\textwidth]{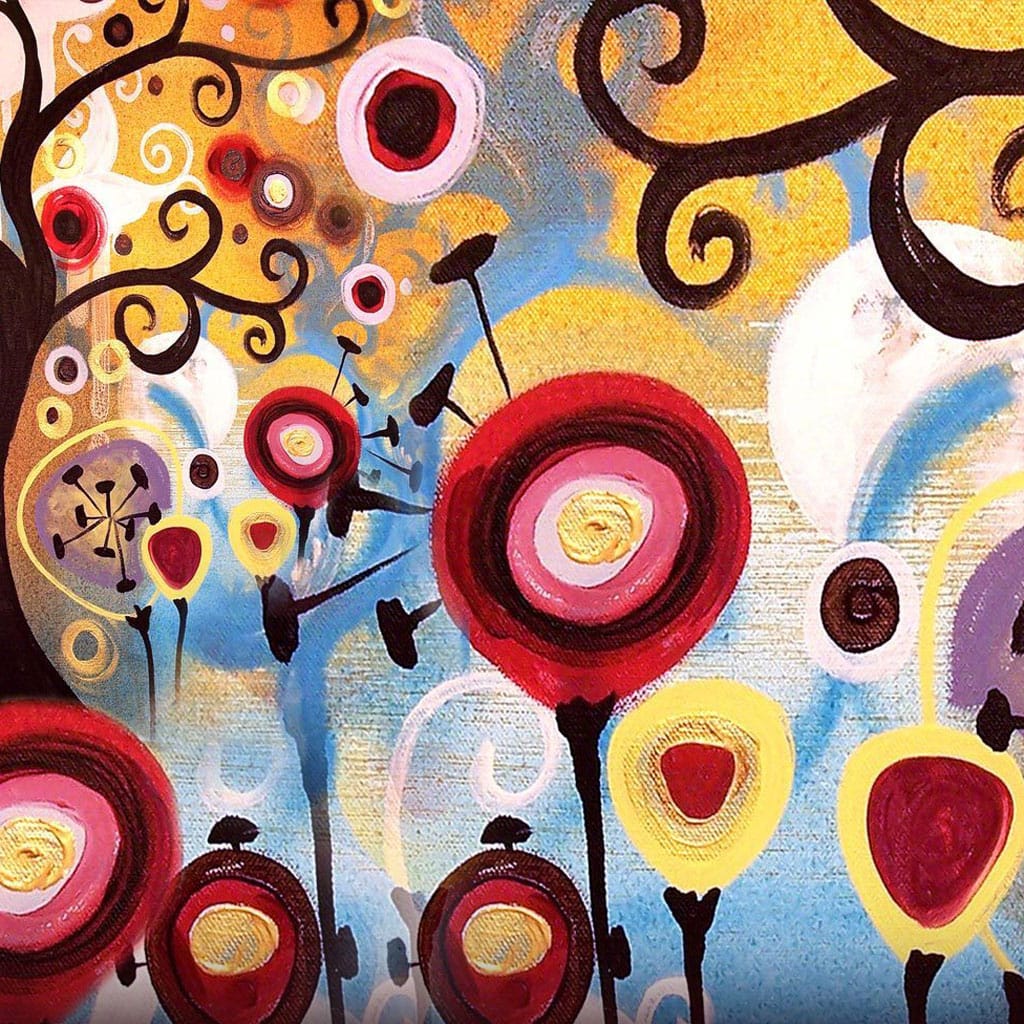}
    \end{subfigure}
    \\
    \begin{subfigure}[b]{0.30\textwidth}
        \includegraphics[width=\textwidth]{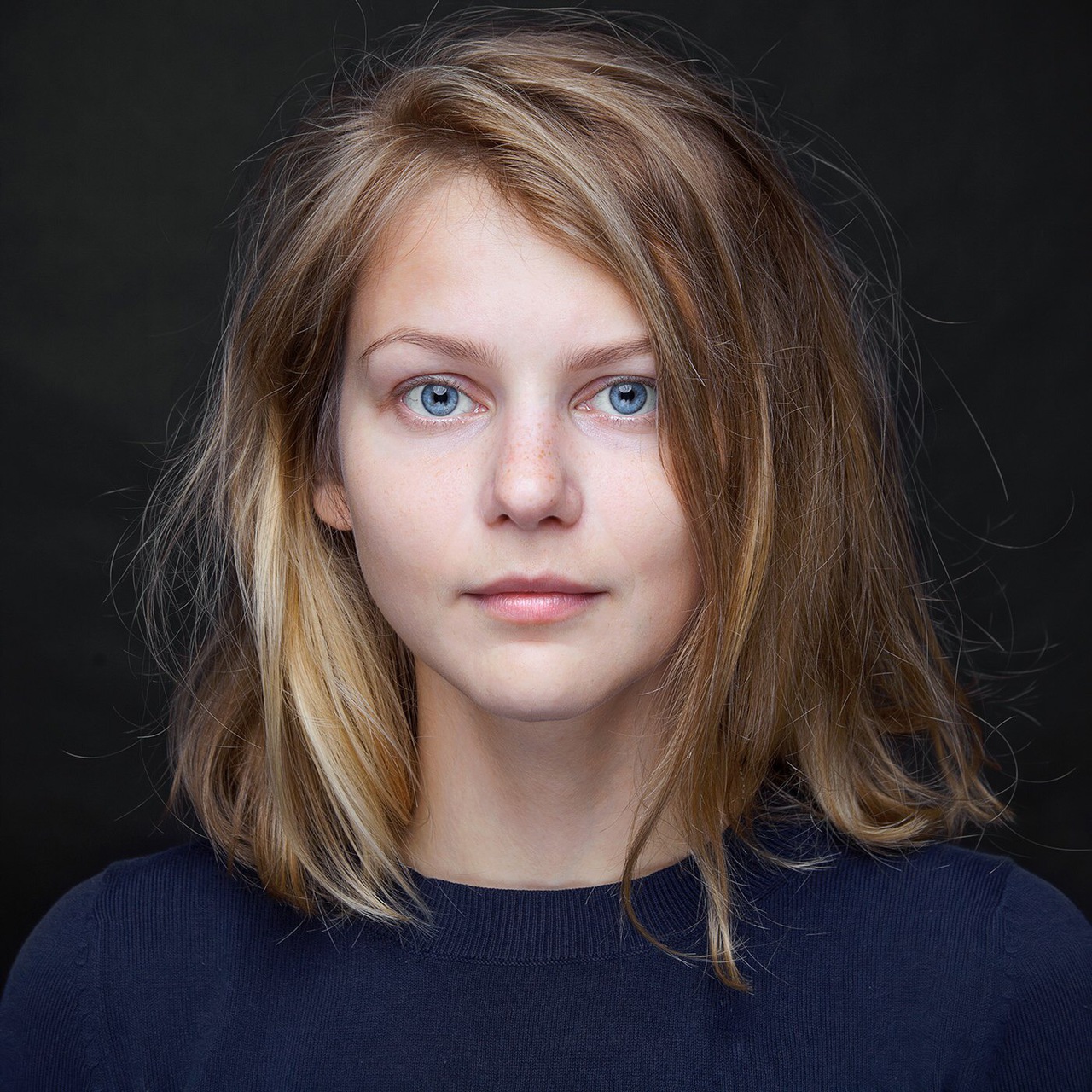}
    \end{subfigure}
    \begin{subfigure}[b]{0.30\textwidth}
        \includegraphics[width=\textwidth]{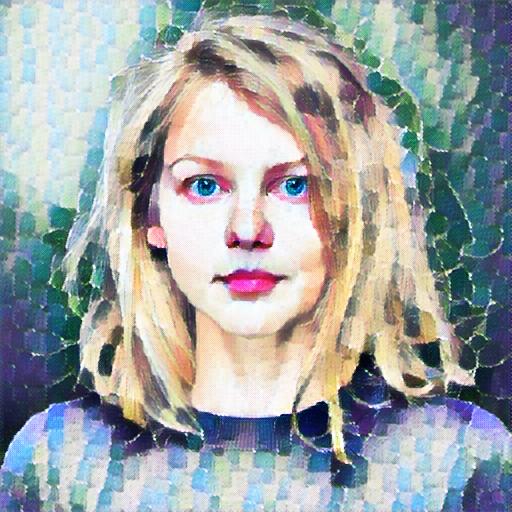}
    \end{subfigure}
    \begin{subfigure}[b]{0.30\textwidth}
        \includegraphics[width=\textwidth]{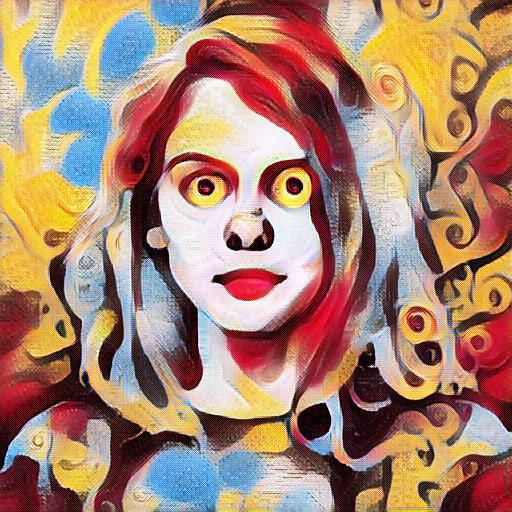}
    \end{subfigure}
\caption{Stylization examples using proposed method. First row: style images; second row: original image and its stylized versions.}\label{fig:results}
\end{figure}

Next, the replaced batch normalization with instance normalization and retrained the generators using the same hyperparameters. We found that both architectures significantly improved by the use of instance normalization (\cref{fig:before_after} second row). The quality of both generators is similar, but we found the residuals architecture of \cite{DBLP:journals/corr/JohnsonAL16} to be somewhat more efficient and easy to use, so we adopted it for the results shown in \cref{fig:results}.   

\begin{figure}
    \centering
    \begin{subfigure}[b]{0.45\textwidth}
        \includegraphics[width=\textwidth]{figures/results/karya512.jpg}
      \end{subfigure}
    \begin{subfigure}[b]{0.45\textwidth}
        \includegraphics[width=\textwidth]{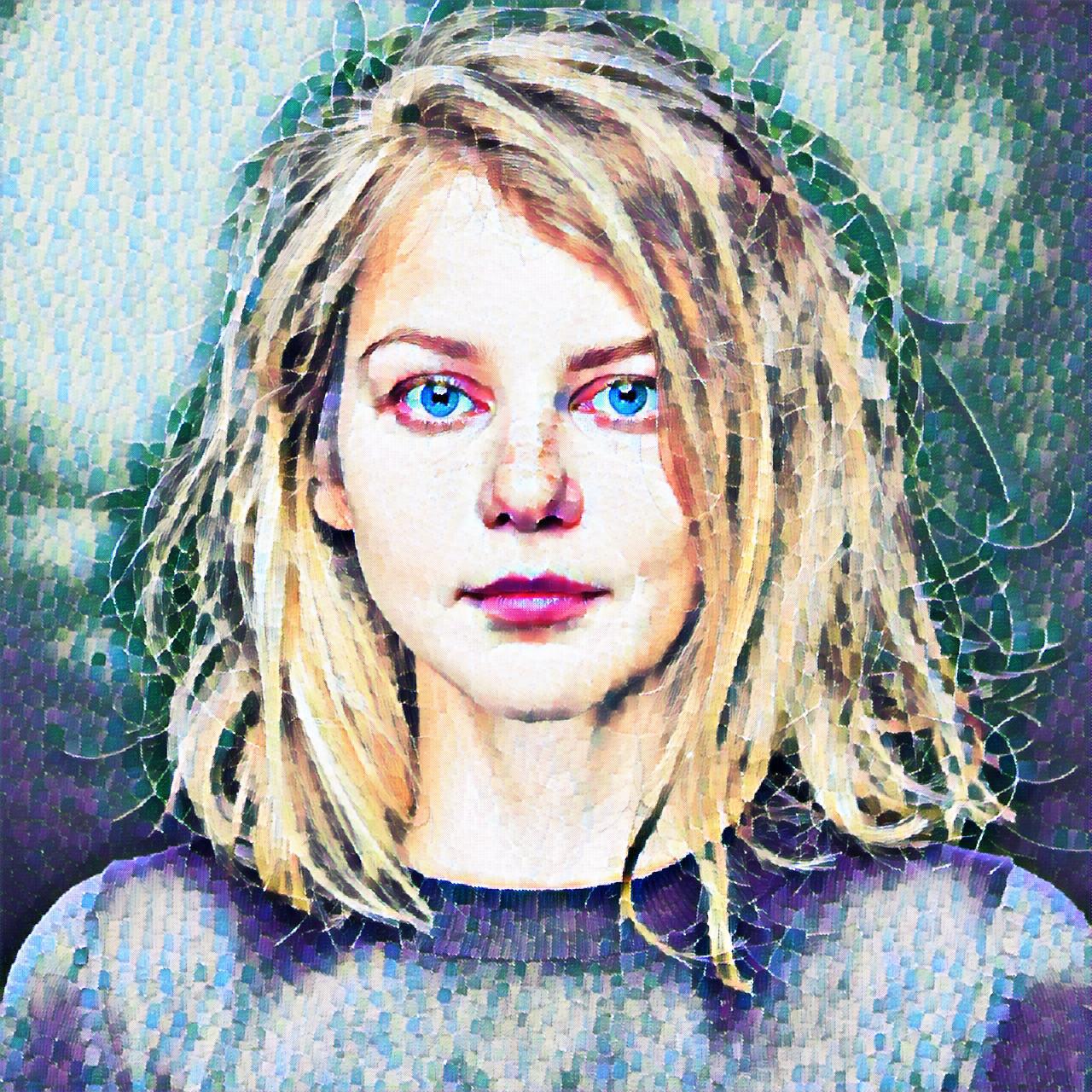}
    \end{subfigure}
\caption{Processing a content image from \cref{fig:results} with Delaunay style at different resolutions: 512 (left) and 1080 (right).}
\end{figure}

\section{Conclusion}\label{s:conc}

In this short note, we demonstrate that by replacing batch normalization with instance normalization it is possible to dramatically improve the performance of certain deep neural networks for image generation. The result is suggestive, and we are currently experimenting with similar ideas for image discrimination tasks as well.

\bibliographystyle{apalike}
\bibliography{main}

\begin{thebibliography}{}

\bibitem[Gatys et~al., 2016]{Gatys_2016_CVPR}
Gatys, L.~A., Ecker, A.~S., and Bethge, M. (2016).
\newblock Image style transfer using convolutional neural networks.
\newblock In {\em The IEEE Conference on Computer Vision and Pattern
  Recognition (CVPR)}.

\bibitem[Johnson et~al., 2016]{DBLP:journals/corr/JohnsonAL16}
Johnson, J., Alahi, A., and Li, F. (2016).
\newblock Perceptual losses for real-time style transfer and super-resolution.
\newblock {\em CoRR}, abs/1603.08155.

\bibitem[Ulyanov et~al., 2016]{DBLP:conf/icml/UlyanovLVL16}
Ulyanov, D., Lebedev, V., Vedaldi, A., and Lempitsky, V.~S. (2016).
\newblock Texture networks: Feed-forward synthesis of textures and stylized
  images.
\newblock In {\em Proceedings of the 33nd International Conference on Machine
  Learning, {ICML} 2016, New York City, NY, USA, June 19-24, 2016}, pages
  1349--1357.

\end{thebibliography}
\end{document}